\documentclass[
]{ceurart}

\usepackage{listings}
\newcommand{\name}{\textsc{BioTabQA}}

\begin{document}

\copyrightyear{2022}
\copyrightclause{Copyright for this paper by its authors.
  Use permitted under Creative Commons License Attribution 4.0
  International (CC BY 4.0).}

\conference{CLEF 2022: Conference and Labs of the Evaluation Forum, 
    September 5--8, 2022, Bologna, Italy}


\title{BioTABQA: Instruction Learning for Biomedical Table Question Answering}

\author{Man Luo}[%
email=mluo26@asu.edu, 
]
\author{Sharad Saxena}[%
email=ssaxen18@asu.edu, 
]
\author{Swaroop Mishra}[%
email=smishr1@asu.edu, 
]
\author{Mihir Parmar}[%
email=mparmar3@asu.edu, 
]
\author{Chitta Baral}[%
email=chitta@asu.edu, 
]
\address{Arizona State University, Tempe, Arizona, 85281, United State}



\begin{abstract}
  Table Question Answering (TQA) is an important but under-explored task. 
Most of the existing QA datasets are in unstructured text format and only few of them use tables as the context.
To the best of our knowledge, none of TQA datasets exist in the biomedical domain where tables are frequently used to present information. 
In this paper, we first curate a table question answering dataset, \name, using 22 templates and the context from a biomedical textbook on differential diagnosis. 
\name{} can not only be used to teach a model how to answer questions from tables but also evaluate how a model generalizes to unseen questions, an important scenario for biomedical applications.  
To achieve the generalization evaluation, we divide the templates into 17 training and 5 cross-task evaluations. 
Then, we develop two baselines using single and multi-tasks learning on \name.
Furthermore, we explore instructional learning, a recent technique showing impressive generalizing performance.
Experimental results show that our instruction-tuned model outperforms single and multi task baselines on an average by {$\sim23\%$} and {$\sim6\%$} across various evaluation settings, and more importantly, instruction-tuned model  outperforms baselines by $\sim5\%$ on cross-tasks.
\end{abstract}


\maketitle

\section{Introduction}


Neural language models have achieved state-of-the-art performance in popular reading comprehension (RC) tasks such as SQuAD \cite{rajpurkar2016squad, rajpurkar2018know}, DROP \cite{dua2019drop} and ROPES \cite{lin2019reasoning}. Unlike in popular RC where the context contains information in natural language, 
a significant amount of real-world information 
is stored in unstructured or semi-structured web tables \cite{chen2020open}. In particular, many clinical information is provided in tabular format \cite{durbin2004effective}. Past attempts have been made for TQA in the general domain Natural Language Processing (NLP) \cite{jauhar2016tables, pasupat2015compositional, iyyer2017search}, however, this task has not been well-studied in the biomedical domain. 

This work takes the first step toward studying the TQA task in biomedical domain. To this extent, we first curate a table question answering dataset, \name, using 22 templates without heavy and expensive human annotation. 
This dataset also serves to evaluate the generalization of a model, a well-known issue that many language models have failed even though they outperform humans in many popular benchmarks ~\cite{eykholt2018robust, le2020adversarial}. Recently, instruction-learning \cite{mishra2021cross, wei2021finetuned, parmar2022boxbart} have improved model's performance to unseen tasks. 
Inspired by this, we leverage instruction-tuning to build a model and verify whether instruction learning also show stronger generalization on \name.



Our contribution can be summarized as: (1) to the best of our knowledge, this is the first attempt to study biomedical TQA and this is also the first attempt to incorporate instructional learning in this task, (2) we reformulate differential diagnosis as a TQA problem and introduce a new dataset \name, and (3) experimental results show that our instruction-tuned model outperforms single and multi task baselines by 23\%, 6\%,  and outperforms  multitask model by 5\% in cross-task (generalization to unseen task) setting. Finally, our analysis shows that instruction is more important and useful in cross tasks compared to in-domain tasks in inference time. 


\section{Related Work}
\label{app:related_work}

\paragraph{Table Question Answering} Past attempts have been made for TQA such as TabMCQ \cite{jauhar2016tables}, WikiTableQuestions \cite{pasupat2015compositional}, Sequential Q\&A \cite{iyyer2017search}, Spider \cite{yu2018spider}, WikiSQL \cite{zhong2017seq2sql}. These approaches can handle the large-scale tables from Wikipedia efficiently. However, these QA systems can only answer the question when a strong signal needed for identifying the type of answers is provided explicitly in the table. To overcome this limitation, TabFact \cite{chen2019tabfact} is proposed which enables TQA when the answer is not explicitly available in the table. 
However, none of the above datasets are in the biomedical domain, a domain which is not only essential in human life but also in which tables have wide applications (e.g. many biomedical information are presented by tables). 
There exists some table datasets in biomedical domain, such as PubTabNet \cite{zhong2020image},  a medical table datasets which are widely used in information retrieval tasks.
Some other datasets are designed for biomedical question answering task such as \cite{tsatsaronis2012bioasq, jin2019pubmedqa}.
Nevertheless, these are not biomedical TQA dataset, leaving the biomedical TQA as an under-explored task.
This work aims to take the first step to Table question answering in biomedical domain, which is known to be different from the general domain~\cite{lee2020biobert,luo2022improving}.

\paragraph{Instruction Learning} Recently, the paradigm in ML/DL shifted to prompt-based learning. \citet{liu2021pre} provides a comprehensive survey on prompt-based methods for various tasks. Prompts enable the generalization across tasks as well as achieves considerable performance on zero-shot learning. T0 model \citet{sanh2021multitask} shows effective performance on multi-tasking and zero-task generalization using a prompt-based approach. \cite{mishra2021cross} introduced natural language instructions to improve the performance of LMs such as BART, GPT-3 for cross-task. Followed by this, FLAN \cite{wei2021finetuned} has been proposed which uses instructions to achieve generalization across unseen tasks. Recently, \citet{parmar2022boxbart} proposed instruction learning for biomedical multi-task. Along with that, \citet{mishra2021reframing} shows reframing instructional prompts can boost both few-shot and zero-shot model performance. \citet{min2021metaicl} shows performance of in-context learning on a large set of training tasks. InstructGPT model is proposed which is fine-tuned with human feedback \cite{ouyang2022training}. Instruction-based multi-task framework for few-shot Named Entity Recognition (NER) has been developed by \citet{wang2022instructionner}. \citet{puri2022many} introduced instruction augmentation and \citet{prasad2022grips} introduced Gradient-free Instructional Prompt Search (GrIPS) for improving model performance. Recently, \citet{parmar2022don} believe that instruction bias in existing Natural Language Understanding (NLU) datasets can impact the instruction learning, however, many approaches have been proposed recently using instructions to improve model performance \cite{wu2021ai, wu2022promptchainer, lin2021few, kuznia2022less, wang2022benchmarking}. Motivated by the effectiveness of instruction learning, in this work, we explore the potential application of instructional prompts for the biomedical TQA.
\begin{table*}[t]
\centering
\resizebox{0.95\linewidth}{!}{
\begin{tabular}{p{0.5cm}p{7.5cm}p{7.5cm}}
\toprule
 \textbf{ID} & \textbf{Question Template} &\textbf{ Prompt}\\ 
 \midrule
 1 & I have symptom A, what disease do I have? & If symptom A is in symptom list, report corresponding disease. \\   
 \midrule
 2 & I have symptom A and sign A, what is my diagnosis? & If symptom A is in symptom list, and sign A is in sign list, report corresponding disease.\\
 \midrule
 11 & The patient has symptom A, symptom B and symptom C, what disease can cause these symptoms? & If symptom A, symptom B and symptom C are in symptom list, report corresponding disease.\\
 \midrule
 22 & I have symptom A, symptom B, symptom C but no symptom D, what is causing this?& If symptom A, symptom B and symptom C are in symptom list, but symptom D is not in symptom list, report corresponding disease.\\
 \bottomrule
 \end{tabular}
 }
 \caption{Examples of four question templates for BioTabQA dataset creation and the corresponding prompts}
 \label{tab:temp_prompt}
\end{table*}

\section{Task and Dataset}\label{sec:dataset}

\paragraph{Task Formulation}
Each data point is a tuple <T, Q, A>, where T is a table, Q is a question, and A is the answer to Q in T. 
In particular, Q exhibits some symptoms/signs and asks about what potential disease it is, e.g., ``I have joint pain and swelling on my face, what's wrong with me''. A is the corresponding disease (or diagnosis) in T. 
The task is to predict A given <T, Q> as input. 

\paragraph{Dataset Source} 
We use the medical textbook ``Differential Diagnosis in Primary Care''~\cite{rasul2009differential} as the source of our dataset, which contains information on how to diagnose a patient by observing their disease symptoms.  
This book is in the tabular format with five columns: (1) diagnosis, (2) key symptoms, (3) key signs\footnote{According to JAMA Network, a symptom is a manifestation of disease appears to the patient himself, while a sign is a manifestation of the disease that the physician perceives.}, (4) background, and (5) additional information. We only use the first three columns to create the dataset. We divide the textbook into 513 tables.

\paragraph{Dataset Creation}
To create large scale training/evaluation datasets (i.e., \name) without laborious human annotation, we design a wide range of templates to semi-automate the process of dataset generation. 
We use key symptoms and/or key signs of a diagnosis in the question templates, and the diagnosis as the answer. 
In addition, we design the corresponding prompt to enable instruction learning for each template. 
In total, we design 22 templates. Table \ref{tab:temp_prompt} shows four templates as an example and the corresponding prompts (all templates and prompts are given in Appendix \ref{app:biotabqa}.
Specifically, some templates have one, two, three or four symptoms/sign (e.g. ID 1, 2, 11, 22), and some have negation (e.g. ID 22). 
Once the templates are pre-defined, given a table, and a template, for each row, we randomly select the symptoms/signs based on the template and replace the placeholder in the template with the chosen symptoms/signs.

\paragraph{Three Splits in \name}
For experimental purposes, we created 3 training/testing/cross-task splits of data. 
Each split includes 17 templates for in-domain training and testing, where there are non-overlap tables for training and testing. 
The rest 5 templates are used for cross-task evaluation. 
For each split, the templates are similar to each other in the training set and less similar to the templates in the evaluation set (cross-task setting) to show the generalization capability of a model. 
The similarity is defined as either the same number of symptoms/signs presented in the templates or similar phrases in the templates.
Table \ref{tab:split1} shows the statistics of Split 1 and other Splits as well as the division of the Splits are given in Appendix \ref{app:biotabqa}.
\begin{table}[t]
\centering
\resizebox{0.8\linewidth}{!}{
\begin{tabular}{cccc}
\toprule
Statistic  & Train   &  IID Test & Cross Task Test\\ 
\midrule
\# of Samples & 9,126   &  19,590 & 2,463 \\
Question Length & 240   &  20 & 16 \\
Table Length  & 255   &  256 & 246 \\
Prompt Length & 18   &  18 & 14 \\
\# Tasks with 1 sym/sign & 0   &  0 & 3 \\
\# Tasks with 2 sym/sign & 9   &  9 & 2\\
\# Tasks with 3 sym/sign & 7   &  7 & 0\\
\# Tasks with 4 sym/sign & 1   &  1 & 0\\
\# Tasks with negation & 2   &  2 & 0\\
\hline
\end{tabular}
}
\caption{Statistic of BioTabQA Split 1 for training (Train), in-domain testing (IID Test) and cross task testing (Cross Task Test) sets.}
\label{tab:split1}
\end{table}


\section{Experiments and Results}

From our dataset, each question type (i.e., template) is considered an individual task. Hence, we have 22 different tasks in total. 
We design two baselines, the single-task model (STM) and the multi-tasks model (MTM).
We compare the performance of the instruction-tuned model (In-MTM) with these two baselines on the in-domain test set, cross-task, and robustness~\cite{kitano2004biological,gokhale2022generalized}. 
We use DistilBert~\cite{sanh2019distilbert} as the backbone model for all experiments. Exact Match (EM) score is used as an evaluation metric. Other experimental setup can be found in Appendix \ref{app:experimental_setup}. 
In the following, we describe the table linearization technique followed by our instructional multi-task learning model. We present the results and analysis at the end of this section.  

\subsection{Table Linearization}
Since input of the language model is text, we need to linearize the table context from BioTabQA.
We use a simple yet effective linearization method suggested by \cite{chen2019tabfact} to convert the table context into a string of text.
We pre-define the format ``Row 1 is: Diagnosis is \_, Key symptoms are \_, Key signs are \_;..., Row N is Diagnosis is \_, Key symptoms are X, Key signs are XX''. 

\subsection{Instructional Multi-task Learning Model}
Apart from the prompt designed for each template (see \S\ref{sec:dataset}), one additional example is also given in the instruction. 
The example consists of a question and the answer without the context table due to the input length restriction of the language model. 
We also use special words to denote the beginning of the prompt, and question and answer. In particular, the instruction set of \{Prompt: p. Question: q. Answer: a\}. The input to our instruction learning model is \{[CLS] Question: Q, Context: C, Instruction: I\}, where [CLS] is the special token of the DistilBERT model, Q is the input question, C is the input table after linearization. 
As mention in \S\ref{sec:dataset}, we create multiple templates and we term the data created by individual template as task. A single task model (STM) is trained by one task, and a multitask model (MTM) is trained by multiple tasks. 

\subsection{Main Results} 

We evaluated our proposed model In-MTM in terms of various aspects including in-domain testing, cross-task setting and robustness. All the results are presented in Table \ref{tab:main_result_split}.
In the following, we present insightful results and findings based on our experiments. 
The performance of MTM and In-MTM varies for different split since each split consists of different tasks.

\begin{table*}[t]
\resizebox{\linewidth}{!}{
\begin{tabular}{ccccccccc}
\toprule
\multirow{2}{*}{\textbf{Task ID}} & \multirow{2}{*}{\textbf{\# Training}}   & \multirow{2}{*}{\textbf{STM}}  & \multicolumn{2}{c}{\textbf{Split 1}}  & \multicolumn{2}{c}{\textbf{Split 2}}& \multicolumn{2}{c}{\textbf{Split 3}}\\
~ & ~ &~ &  \textbf{MTM} &  \textbf{In-MTM} &  \textbf{MTM} &  \textbf{In-MTM} &  \textbf{MTM} &  \textbf{In-MTM} \\ \hline
1 & 667 & 0.53 &  \colorbox{green!30}{0.84} &  \colorbox{green!30}{0.85} &  0.88  &  \textbf{0.91}  & \colorbox{green!30}{0.88} & \colorbox{green!30}{0.88}  \\ 
2 & 3023 & 0.55  & {0.83} & {0.93} & 0.88 & \textbf{0.94} & 0.87 & 0.93 \\
3 & 3082 & 0.62  & {0.87} & {0.93 }& 0.89 & \textbf{0.96} & \colorbox{green!30}{0.90} & \colorbox{green!30}{0.95}  \\
4 & 3170 & 0.64 &  \colorbox{green!30}{0.80} &  \colorbox{green!30}{0.92} &  0.88 &  0.93  &  0.87 &  \textbf{0.94}  \\ 
5 & 47561 & 0.90  & 0.88 & {0.93} & 0.92 & \textbf{0.95}  & 0.90 & 0.93 \\
6 & 10991  & 0.86  & 0.86 & {0.94} & 0.89 & \textbf{0.98} & 0.91 & 0.96  \\
7 & 3082 & 0.60  &  \colorbox{green!30}{0.87} &  \colorbox{green!30}{0.93} &  0.89 &  \textbf{0.97} &  0.89 &  0.95 \\ 
8 &3082  & 0.60  & {0.87}  & {0.92}  & \colorbox{green!30}{0.89} & \colorbox{green!30}{0.92} & 0.89 & \textbf{0.95} \\
9 & 10324 & 0.82  & 0.80     & {0.95}  & \colorbox{green!30}{0.83} & \colorbox{green!30}{\textbf{0.96}}  & 0.83 & \textbf{0.96}  \\
10 & 3082 & 0.63  & {0.87}    & {0.93 } & 0.89 & \textbf{0.96} & 0.90 & 0.95 \\
11 & 10991& 0.88  & 0.86 & {0.94} & \colorbox{green!30}{0.89} & \colorbox{green!30}{\textbf{0.97}}  & 0.91 & 0.95  \\
12 & 3082 & 0.60  & {0.87}  & {0.92  } & \colorbox{green!30}{0.89} & \colorbox{green!30}{\textbf{0.95}}   & 0.89 & 0.94 \\
13 & 10991 & 0.90  & 0.86   & {0.93}  &  0.89& \textbf{0.98} & 0.90 & {0.95} \\
14 &10991 & 0.71  & {0.86}    & {0.93} & \colorbox{green!30}{0.89} & \colorbox{green!30}{\textbf{0.98}}  & 0.90 & 0.96 \\
15 & 667 & 0.51 &  \colorbox{green!30}{0.84} &  \colorbox{green!30}{0.85} &  0.89 &  \textbf{0.90} & \colorbox{green!30}{0.87} & \colorbox{green!30}{\textbf{0.90}}  \\ 
16 & 3082 & 0.63  & {0.87}      & {0.92} & 0.89 & \textbf{0.96} & \colorbox{green!30}{0.89} & \colorbox{green!30}{0.92}  \\
17 & 10991 & 0.80  & {0.87}    & {0.94}  & 0.89 & \textbf{0.98} & \colorbox{green!30}{0.90} & \colorbox{green!30}{0.95}  \\
18 & 3082 & 0.68  & {0.88}    & {0.93}   &0.89 &\textbf{0.96} & 0.89 & 0.94 \\
19 & 3082 & 0.60  & {0.87}    & {0.93}  & 0.89 & \textbf{0.96} & 0.90 & 0.95  \\
20 &3082 & 0.61  & {0.87}    & \textbf{0.91}  & 0.89 &\textbf{0.95}  & 0.90 & 0.94 \\
21 & 667 & 0.54  &  \colorbox{green!30}{0.83} &  \colorbox{green!30}{0.85} &  0.87 &  0.89  &  0.85& \textbf{0.90}\\
22 & 14639 & 0.88  & {0.89}    & {0.93}   & 0.93 & \textbf{0.97 }& 0.92 & 0.94 \\ \hline
Avg. Split 1  & 9127 & 0.72 & {0.86}& \textbf{0.93}  \\
Avg. Split 2  & 7349 & 0.68 & - & -  & 0.89  & \textbf{0.95} \\
Avg. Split 3  & 8525 & 0.71  & - & - &-&-& 0.89 &\textbf{0.94} \\
Avg. cross Split 1  & - & - & {0.84}& \textbf{0.88}  \\
Avg. cross Split 2  & - & - & - & - & 0.88 & \textbf{0.97} & - & - \\
Avg. cross Split 3  & - & - & - & - &- & - &0.89 & \textbf{0.92} \\
\hline
\end{tabular}
}
\caption{The EM (exact matching) scores of three Models on BioTabQA. \colorbox{green!30}{Green} denotes cross task performance. \textbf{Bold number} denotes the best performance for each task.}
\label{tab:main_result_split}
\end{table*}

\begin{figure}[t]
    \centering
    \includegraphics[width=0.85\linewidth]{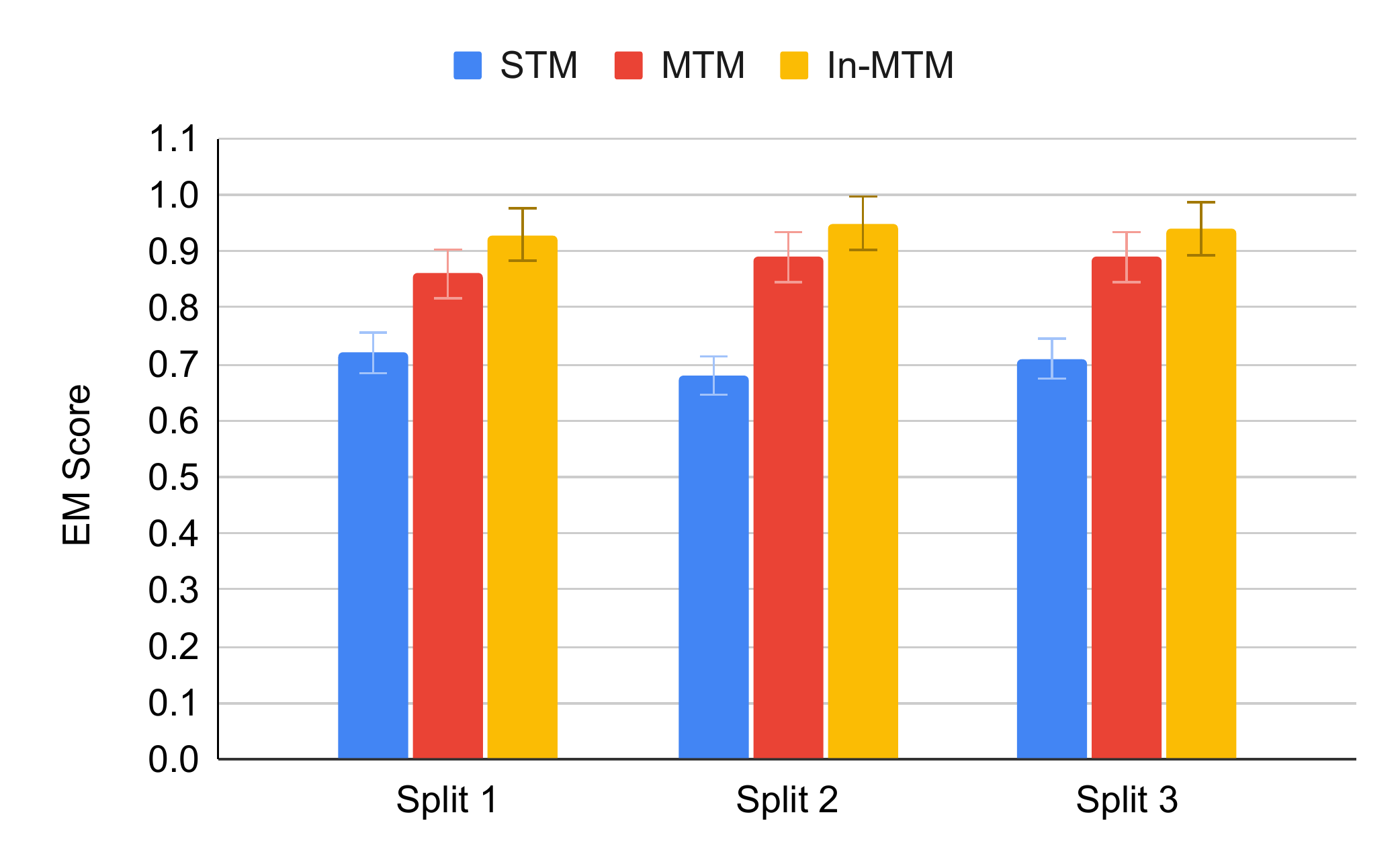}
    \caption{The average performance of three models on the in-domain testing sets of different Splits.}
    \label{fig:multi_task_results}
\end{figure}

\paragraph{Finding 1: Multitask Model performs better than Single-task Model} 
From Figure \ref{fig:multi_task_results}, we can observe that MTM outperforms STM in majority cases, leading to on an average 14\%, 21\%, and 18\% improvement on split 1, 2, and 3, respectively. Also, we observe that multi-task learning is significantly helpful on the tasks where the training data is less. Hence, we observe tasks 1, 15, and 21 which have only 667 training examples (see results in the first block of Table \ref{tab:finding1_result_split}). We can see that the STM model achieve less than 0.60 EM score; while the MTM trained on split 2 achieves at least 0.85 EM score\footnote{We compare STM with MTM only on split 2 results in this scenario because task 1, 15, and 21 are used for training in split 2.}. For split 1 where the MTM does not train on tasks 1, 15 and 21 tasks, it shows superior performance compared to the STM. This indicates that multi-task learning is effective in a low-resource setting for TQA. Moreover, for tasks 5 and 13 which have more than $10k$ instances, the STM can obtain a 0.90 EM score; while the MTM trained on the split 2 achieves similar performance. This finding is aligned with the literature that multitask learning model improves single task learning model~\cite{mccann2018natural,fisch2019mrqa,luo2022choose}

\begin{table}[t]
\resizebox{0.8\linewidth}{!}{
\begin{tabular}{cccccc}
\toprule
\multirow{2}{*}{\textbf{Task ID}} & \multirow{2}{*}{\textbf{STM}}  & \multicolumn{2}{c}{\textbf{Split 1}}  & \multicolumn{2}{c}{\textbf{Split 2}} \\
\cmidrule(lr){3-4} \cmidrule(lr){5-6}
~ & ~ &  \textbf{MTM} &  \textbf{In-MTM} &  \textbf{MTM} &  \textbf{In-MTM} \\
\hline
1 & 0.53 &  0.84 &  0.85 &  0.88  &  0.91 \\ 
15 & 0.51 &  0.84 &  0.85 &  0.89 &  0.90 \\ 
21 & 0.54  &  0.83 &  0.85 &  0.87 &  0.89 \\
\hline
5 & 0.90  & 0.88 & 0.93 & 0.92 & 0.95  \\
13 & 0.90  & 0.86   & 0.93 &  0.89& 0.98 \\
\bottomrule
\end{tabular}
}
\caption{The EM (exact matching) scores of STM, MTM and In-MTM on the low resources tasks (first block) and high resources tasks (second block).}
\label{tab:finding1_result_split}
\end{table}


\paragraph{Finding 2: Instruction further improve Multi-task learning Model} 
We can observe from Figure \ref{fig:multi_task_results} that In-MTM further improves the performance of the MTM, yielding on an average 6\%, 6\% and 5\% improvement on split 1,2, and 3, respectively. These results indicate the use of instructional prompts increases the question-answering performance both consistently and significantly.



\begin{table*}[t]
\center
\resizebox{0.95\linewidth}{!}{
\begin{tabular}{cccccc}
\toprule
\textbf{Task ID}    &  \textbf{Correct } & \textbf{Mismatched  } & \textbf{Repeat  } & \textbf{Random String  } & \textbf{Random Words }\\ 
\midrule
2  & 0.93 & 0.92 & 0.89 & 0.90 & 0.89 \\
4 & 0.92 & 0.91  & 0.85 & 0.86 & 0.85 \\
\hline
20  & 0.91 & 0.92 & 0.93 & 0.94  & 0.93  \\
22  & 0.93 & 0.95 & 0.96 & 0.95 & 0.96 \\
\hline
\end{tabular}
}
\caption{Test the performance of In-MTM (trained on Split 1) using 4 variants of instructions) on two cross tasks (first block) and two in-domain tasks (second block). }
\label{tab:mismatch}
\end{table*}

\begin{figure}[ht]
    \centering
    \includegraphics[width=0.85\linewidth]{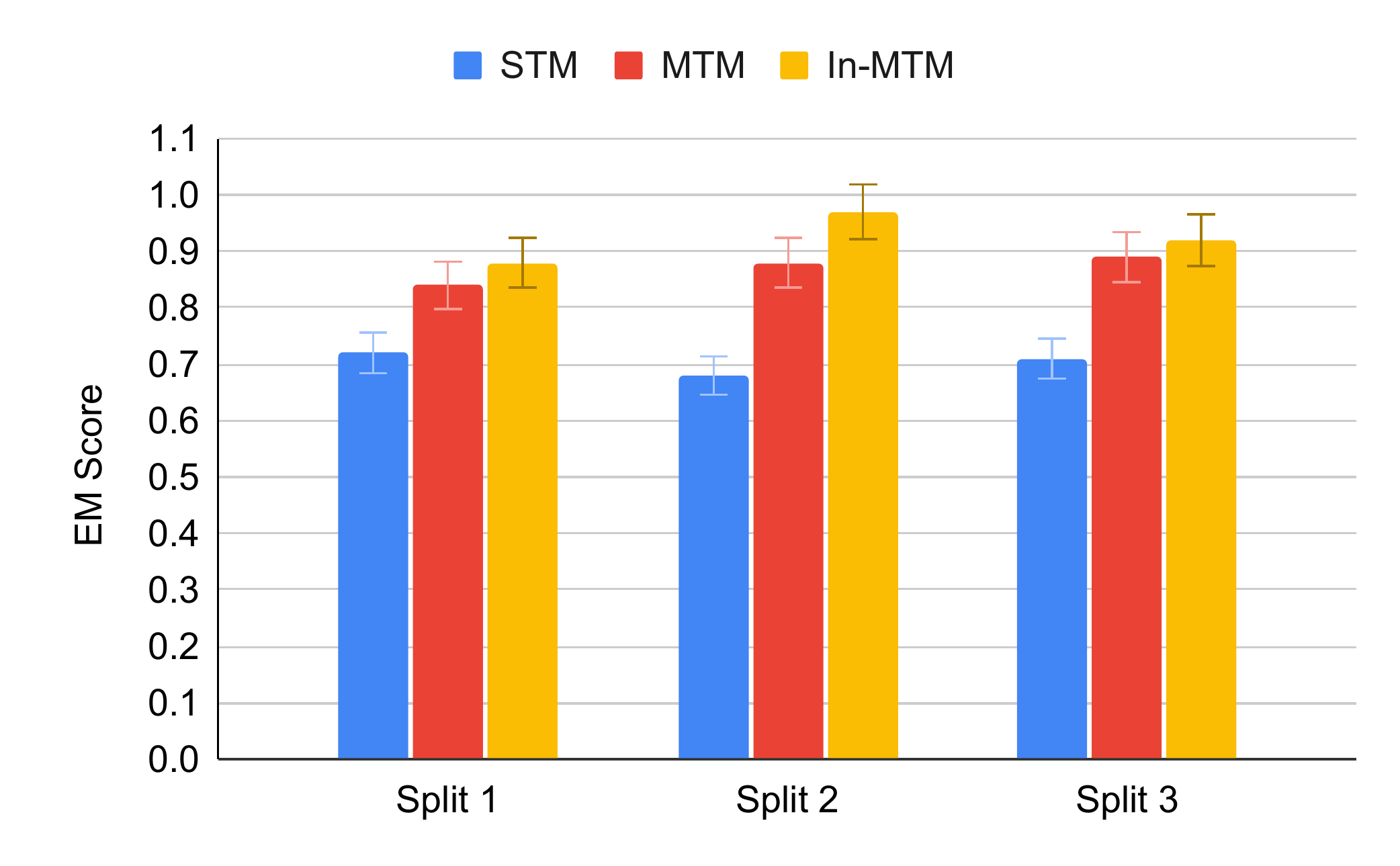}
    \caption{The average performance of three models on the cross-tasks of different Splits.}
    \label{fig:cross_task_results}
\end{figure}

\paragraph{Finding 3: Multi-task learning and Instruction Learning improve generalization capacity of model}
For each split, we hold out 5 different tasks for the cross-task evaluation. This is similar to out-of-domain evaluation where a model has not seen such types of questions in the training time, thus the performance on the cross tasks demonstrates the generalization capacity of a model. From the results shown in Figure \ref{fig:cross_task_results}, we have two observations. 
First, we can see that both MTM and In-MTM show sufficient performance. In split 3, In-MTM achieves average 0.94 EM on in-domain tasks (see Figure \ref{fig:multi_task_results}) and average 0.92 EM on cross-tasks (see Figure \ref{fig:cross_task_results}), a marginal drop ($\sim2\%$). On the same split, MTM achieves the same performance on in-domain tasks and cross-tasks. More importantly, both MTM and In-MTM achieve higher performance than the STM on every task even though the former two models do not train on these tasks.  This demonstrates the benefits of multi-task learning. Second, for each Split, In-MTM achieves better performance than MTM on every cross-task. This shows that instruction learning can further improve generalization.

\paragraph{Finding 4: Instruction is more useful in cross tasks compared to in-domain tasks.}

We evaluate the In-MTM on two cross tasks and in-domain tasks with different types of instructions to analyze the change in model performance. 
First, we use mismatched instruction, i.e., instruction of one task for other tasks, however, it is still instructive. 
From Table \ref{tab:mismatch}, we can see that model shows similar performance with the mismatch instruction as original instruction on both in-domain and cross tasks. 
The reason might be that instructions for the different tasks have a set of similar words (see Appendix \ref{app:biotabqa}), and previous studies~\cite{webson2021prompt, schick2021true} have shown that the model can perform well if the instructions have similar words.
Moreover, we also construct three types of meaningless instructions which does not have any linguistic meaning such as random strings (e.g., `ashlksadkl'), random words (e.g., `hello bye you east'), and repeated characters (e.g., `AAAAA'). 
These meaningless instructions hamper the performance of cross tasks more significantly than in-domain tasks. 
For in-domain tasks, the model is already exposed to the same type of instances at training time, but for cross-task, these unseen tasks heavily rely on the instructions \cite{mishra2021cross}. 
In summary, instructions can be more important and helpful in cross-task settings.
\section{Future Work and Conclusion}

In this work, we take the first step toward studying table question answering in biomedical domain. 
We firstly create a dataset, \name, based on templates using a primary care textbook.
We then experiment with three models on \name{} and find that multi-task learning is better than single task learning especially in low resource scenarios. 
Furthermore, the instruction learning can significantly improve the model without instructions on both in-domain as well as cross-tasks. 
This suggest the benefits of instruction learning on table question answering task, and explore the role of instruction learning in other general table question answering datasets is one interesting future work. 
The questions in current dataset are based on the formal symptoms or signs given in the textbook, which make some questions unnatural. 
Using more natural terms to generate the question can produce a  dataset more close to real-life scenario.

\bibliography{sample-ceur,custom}

\appendix
\clearpage
\section{Details of BioTabQA}
\label{app:biotabqa}
Table \ref{tab:temp_promts_all} shows the 22 templates and the corresponding prompts for BioTabQA datasets.  Table \ref{tab:dataset_split} shows the three Split division including which tasks in the training and cross-task evaluation. 
Table \ref{tab:split2} and  \ref{tab:split3} show the statistic of Split 1 and 2, respectively.
\begin{table*}[t]
\small
\resizebox{\linewidth}{!}{
\begin{tabular}{p{0.5cm}p{7.5cm}p{7.5cm}}
\toprule
 ID & Question Template & Prompt\\ 
 \midrule
 1 & I have symptom A, what disease do I have? & If symptom A is in symptom list, report corresponding disease. \\   
 \midrule
 2 & I have symptom A and sign A, what is my diagnosis? & If symptom A is in symptom list, and sign A is in sign list, report corresponding disease.\\
 \midrule
 3 & I have symptom A and symptom B, what is wrong with me? & If symptom A and symptom B are in symptom list, report corresponding disease.\\
 \midrule
 4 & I have sign A and sign B, what disease do you think I have? & If sign A is in sign list, and sign B is in sign list, report corresponding disease.\\
 \midrule
 5 & I have symptom A and symptom B but not symptom C, what is my potential diagnosis? & If symptom A and symptom B are in symptom list, but symptom C is not in symptom list, report corresponding disease. \\
 \midrule
 6 & A patient is showing symptom A , symptom B and symptom C, what could be causing this? & If symptom A, symptom B and symptom C are in symptom list, report corresponding disease.\\
 \midrule
 7 & A patient is exhibitng syptom A and sign A, diagnose her & If symptom A is in symptom list, and sign A is in sign list, report corresponding disease.\\
 \midrule
 8 & What disease can cause symptom A and symptom B? & If symptom A and symptom B are in symptom list, report corresponding disease.\\
 \midrule
 9 & What disease causes symptom A, symptom B and sign A? & If symptom A and symptom B are in symptom list, and sign A is in sign list, report corresponding disease.\\  
 \midrule
 10 & If my friend has symptom A and symptom B, then what is his potential diagnosis? & If symptom A and symptom B are in symptom list, report corresponding disease.\\
 \midrule
 11 & The patient has symptom A,symptom B and symptom C, what disease can cause these symptoms? & If symptom A, symptom B and symptom C are in symptom list, report corresponding disease.\\
 \midrule
 12 & Which disease is associated with symptom A and symptom B? & If symptom A and symptom B are in symptom list, report corresponding disease.\\
 \midrule
 13 & A patient is complaining about symptom A, symptom B and symptom C, diagnose him. & If symptom A, symptom B and symptom C are in symptom list, report corresponding disease. \\  
 \midrule
 14 & What disease is responsible for symptom A, symptom B and symptom C? & If symptom A, symptom B and symptom C are in symptom list, report corresponding disease.\\
 \midrule
 15 & I am experiencing symptom A, what is wrong with me? &  If symptom A is in symptom list, report corresponding disease.\\
 \midrule
 16 & Why am I experiencing symptom A and symptom B? & If symptom A and symptom B are in symptom list, report corresponding disease.\\
 \midrule
 17 & I have symptom A, symptom B and symptom C, why is this happening? & If symptom A, symptom B and symptom C are in symptom list, report corresponding disease.\\ 
 \midrule
  18 & A patient is showing symptom A and symptom B, what illness is associated with these symptoms?& If symptom A, symptom B and symptom C are in symptom list, report corresponding disease.\\
 \midrule
 19 & I have symptom A, and symptom B, what disease may I have? & If symptom A and symptom B are in symptom list, report corresponding disease.\\
 \midrule
 20 & I have symptom A and symptom B, what possible disease could I have? & If symptom A and symptom B are in symptom list, report corresponding disease.\\
 \midrule
 21 & What is causing my symptom A? & If symptom A is in symptom list, report corresponding disease.\\  
 \midrule
 22 & I have symptom A, symptom B, symptom C but no symptom D, what is causing this?& If symptom A, symptom B and symptom C are in symptom list, but symptom D is not in symptom list, report corresponding disease.\\
 \bottomrule
 \end{tabular}
 }
 \caption{22 types of templates and the corresponding prompts for BioTabQA datasets.}
 \label{tab:temp_promts_all}
\end{table*}

\begin{table}[t]
\resizebox{0.8\linewidth}{!}{
\begin{tabular}{cccc}
\toprule
Statistic  & Train   &  IID Test & Cross Task Test\\
\midrule
\# of Samples & 7,349   &  15,566 & 16,145 \\
Question Length  & 19  &  21 & 17  \\
Table Length & 239   &  255 & 259 \\
Prompt Length & 17   & 17 & 17  \\
\# Tasks with 1 sym/sign & 3   &  3 & 0 \\
\# Tasks with 2 sym/sign & 9   &  9 & 2\\
\# Tasks with 3 sym/sign & 4   &  4 & 3\\
\# Tasks with 4 sym/sign & 1   &  1 & 0\\
\# Tasks with negation & 2   &  2 & 0\\
\hline
\end{tabular}
}
\caption{Statistic of BioTabQA Split 2 for training (Train), in-domain testing (IID Test) and cross task testing (Cross Task Test) sets.}
\label{tab:split2}
\end{table}
\begin{table}[t]
\resizebox{0.8\linewidth}{!}{
\begin{tabular}{cccc}
\toprule
Statistic  & Train   &  IID Test & Cross Task Test\\
\midrule
\# of Samples & 8,524   &  18,278 & 6924  \\
Question Length  & 240  &  21 & 254 \\
Table Length & 19   &  256 & 18  \\
Prompt Length & 18   &  18  & 14\\
\# Tasks with 1 sym/sign & 1   &  1 & 2 \\
\# Tasks with 2 sym/sign & 9   &  9 & 2\\
\# Tasks with 3 sym/sign & 6   &  6 & 1\\
\# Tasks with 4 sym/sign & 1   &  1 & 0\\
\# Tasks with negation & 2   &  2 & 0\\
\hline
\end{tabular}
}
\caption{Statistic of BioTabQA Split 3 for training (Train), in-domain testing (IID Test) and cross task testing (Cross Task Test) sets.}
\label{tab:split3}
\end{table}

\label{app:splits}

\begin{table*}[t]
\centering
\resizebox{0.85\linewidth}{!}{
\begin{tabular}{ccc}
\hline
 Split & Train/Test &  Cross-Task Test \\ 
 \hline
 1 & 2,3,5,6,8,9,10,11,12,13,14,16,17,18,19,20,22  & 1,4,7,15,21 \\ 
 \hline
 2 & 1,2,3,4,5,6,7,10,13,15,16,17,18,19,20,21,22 & 8,9,11,12,14 \\
 \hline
 3 & 2,4,5,6,7,8,9,10,11,12,13,14,18,19,20,21,22  & 1,3,15,16,17\\
 \hline
 \end{tabular}
 }
 \caption{BioTabQA provides three Splits, each Split has 17 tasks for training, and the rest 5 for cross-task evaluations.}
 \label{tab:dataset_split}
\end{table*}

\section{Experimental Setup}
\label{app:experimental_setup}
We use DistilBERT~\cite{sanh2019distilbert} as the backbone model and load the pretrained model distilbert-base-uncased from Huggingface library~\cite{wolf2020transformers}. All models are optimized by AdamW  with learning rate 5e-5 in 4 epochs, batch size 16. The maximum length input to every model is 512. All models are trained on Tesla V100 machine with one GPU.



\end{document}